\documentclass{article}

\usepackage{arxiv}

\usepackage[utf8]{inputenc}
\usepackage[T1]{fontenc}   
\usepackage{hyperref}       
\usepackage{url}            
\usepackage{booktabs}       
\usepackage{amsfonts}     
\usepackage{nicefrac}     
\usepackage{microtype}      
\usepackage{lipsum}		
\usepackage{graphicx}
\usepackage{doi}
\usepackage{amsmath,amssymb,amsfonts}
\usepackage{array}
\usepackage{multirow}

\title{Hybrid Neural Network-Based Indoor Localisation System for Mobile Robots Using CSI Data in a Robotics Simulator}

\author{ \href{https://orcid.org/0009-0009-1084-1309}{\includegraphics[scale=0.06]{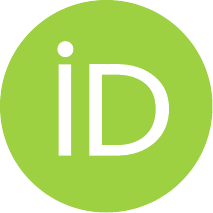}\hspace{1mm}Javier Ballesteros-Jerez} \\
	Computer Systems Department\\
	Universidad de Castilla-La Mancha\\
	Albacete, Spain \\
	\texttt{javier.ballesteros@uclm.es} \\
	\And
    \href{https://orcid.org/0000-0002-4000-1951}{\includegraphics[scale=0.06]{orcid.pdf}\hspace{1mm}Jesus Martínez-Gómez} \\
	Computer Systems Department\\
	Universidad de Castilla-La Mancha\\
	Albacete, Spain \\
	\texttt{jesus.martinez@uclm.es} \\
	\And
    \href{https://orcid.org/0000-0003-3451-7852}{\includegraphics[scale=0.06]{orcid.pdf}\hspace{1mm}Ismael García-Varea} \\
	Computer Systems Department\\
	Universidad de Castilla-La Mancha\\
	Albacete, Spain \\
	\texttt{ismael.garcia@uclm.es} \\
	\And
    \href{https://orcid.org/0000-0003-1510-1608}{\includegraphics[scale=0.06]{orcid.pdf}\hspace{1mm}Luis Orozco-Barbosa} \\
	Computer Systems Department\\
	Universidad de Castilla-La Mancha\\
	Albacete, Spain \\
	\texttt{luis.orozco@uclm.es} \\
    \And
    \href{https://orcid.org/0000-0002-2990-7090}{\includegraphics[scale=0.06]{orcid.pdf}\hspace{1mm}Manuel Castillo-Cara} \\
	Artificial Intelligence Department\\
	Universidad Nacional de Educación a Distancia\\
	Spain \\
	\texttt{manuelcastillo@dia.uned.es} \\
}

\date{}

\renewcommand{\headeright}{}
\renewcommand{\undertitle}{}
\renewcommand{\shorttitle}{Hybrid Neural Network-Based Indoor Localisation System for Mobile Robots}

\hypersetup{
pdftitle={Hybrid Neural Network-Based Indoor Localisation System for Mobile Robots Using CSI Data in a Robotics Simulator},
pdfauthor={Javier Ballesteros-Jerez},
pdfkeywords={Indoor localisation, Positioning, Deep Learning, Hybrid Neural Network, Robotics Simulation},
}

\begin{document}
\maketitle

\begin{abstract}
We present a hybrid neural network model for inferring the position of mobile robots using Channel State Information (CSI) data from a Massive MIMO system. By leveraging an existing CSI dataset, our approach integrates a Convolutional Neural Network (CNN) with a Multilayer Perceptron (MLP) to form a Hybrid Neural Network (HyNN) that estimates 2D robot positions. CSI readings are converted into synthetic images using the TINTO tool. The localisation solution is integrated with a robotics simulator, and the Robot Operating System (ROS), which facilitates its evaluation through heterogeneous test cases, and the adoption of state estimators like Kalman filters. Our contributions illustrate the potential of our HyNN model in achieving precise indoor localisation and navigation for mobile robots in complex environments. The study follows, and proposes, a generalisable procedure applicable beyond the specific use case studied, making it adaptable to different scenarios and datasets.
\end{abstract}


\section{Introduction} \label{sec:1}

Mobile robots are increasingly becoming integrated into various areas of our daily lives, including industrial tasks, medical assistance, and space exploration. Critical to their autonomous operation in these environments is the ability to accurately locate themselves. Without precise localisation, mobile robots face difficulties in basic tasks such as autonomous navigation, operational decision-making, and interaction with objects or people~\cite{mobile_robot}. Therefore, knowing the exact position of the robot at each instant is essential to determine the subsequent actions to be performed.

Recent advancements in robot localisation have significantly improved their autonomous capabilities, particularly through the innovative use of wireless signals. Using technologies such as Wi-Fi, Bluetooth, or ultra-wideband (UWB), researchers have developed more precise and reliable localisation systems, proposing an alternative to traditional methods of robotic localisation~\cite{indoor_wireless, survey_indoor}. Wireless signals enable mobile robots to accurately determine their position in scenarios where traditional methods (e.g. GPS or visual SLAM) are ineffective. By leveraging wireless signals, robots can safely navigate complex indoor spaces, and interact with their surroundings in environments such as warehouses, hospitals, and offices.

In recent years, significant advancements have been made in using Channel State Information (CSI) signals within Massive MIMO (Multiple-Input, Multiple Output) systems for indoor localisation~\cite{mimo_csi}. The combination of CSI with fingerprinting techniques has emerged as a highly effective approach, leveraging the spatial diversity and multipath propagation characteristics inherent in MIMO systems. These advancements are primarily driven by the application of deep learning methodologies~\cite{DL_Fprnt_mimo_csi}, which allows for the direct use of signal readings to determine user locations with high precision, even centimetre-level accuracy in indoor environments~\cite{original}.

Nevertheless, assessing localisation systems in real environments with these techniques presents numerous challenges, including the high costs in infrastructure, time, and human resources required for fingerprinting data collection~\cite{fingerprint_1, fingerprint_2}. To mitigate these problems, existing datasets containing wireless signal readings are useful knowledge sources. In addition, robotics simulators offer an effective solution for evaluating the effectiveness of localisation systems in mobile robots~\cite{rob_simulation}.

The goal of this paper is, therefore, to develop and evaluate robust localisation systems for mobile robots based on wireless CSI readings. Our proposal involves CSI data coming from existing datasets, which are transformed using the TINTO tool~\cite{tinto1, tinto2} to feed a hybrid neural network (HyNN) composed of a Convolutional neural network (CNN) branch with a multilayer perceptron (MLP) branch. The evaluation is performed using a robotics simulator, namely Webots\footnote{\url{https://cyberbotics.com/}}, which eases the design and development of different test cases. Although we have focused on the existing dataset created in~\cite{dataset}, this proposal holds potential for broader applications. The main contributions of this paper are:

\begin{itemize}

\item The development of a HyNN fed from transformed CSI data, which combines a CNN branch with a MLP, and is used to perform robot localisation. 

\item The use of robotics simulators to recreate real scenarios and integrate the localisation solution with the operation of the mobile robot.

\item The integration of the HyNN outputs, namely the robot position, with state estimators (e.g. Kalman filters) to analyse their value in the development of robust and accurate robot localisation systems.

\end{itemize}

The rest of this paper is organised as follows: in Section~\ref{sec:2}, we delve into the related work and review research in indoor localisation using wireless signals, deep learning-based approaches, and robotics simulation tools. Section~\ref{sec:3} presents the proposal and analyses how to develop a localisation model from CSI readings. The experimental design is discussed in Section~\ref{sec:4}, where we also present the obtained results. Finally, Section~\ref{sec:5} outlines the conclusions and future work.

\section{Related Work} \label{sec:2}

\subsection{Measurement of Wireless Signals for Indoor Localisation}

Recent advancements in indoor localisation employ wireless signal measurements, particularly those associated with Bluetooth, Wi-Fi, and UWB technologies~\cite{survey_indoor_2}. Wi-Fi-based localisation is a popular method due to its widespread availability and ease of deployment, typically using Received Signal Strength (RSS). Although RSS methods are simple and cost-effective, they are susceptible to inaccuracies due to environmental variations and signal fluctuations~\cite{rssi}. To overcome these limitations, researchers have explored MIMO systems~\cite{mimo_csi}, which can simultaneously transmit and receive multiple data streams, through the use of multiple transmission and reception antennas. This feature enhances the robustness and reliability of wireless communication, and highlight MIMO systems as a promising tool for accurate localisation.

A significant advancement in indoor localisation is the use of CSI. CSI provides detailed information on signal propagation, capturing amplitude and phase information for each subcarrier, thereby enabling more precise localisation than RSS methods. Represented as a matrix of complex numbers, CSI leverages the spatial diversity of MIMO systems to create a comprehensive environment profile, thus facilitating high-accuracy indoor localisation~\cite{csi}.

\subsection{Fingerprinting}

Fingerprinting is a scene analysis technique that generates environment maps using wireless signal measurements from access points. It involves an offline phase, where reference measurements (e.g., RSS or CSI values) are collected at known locations to build a signal map, and an online phase, where real-time measurements are matched against this map to determine the device position. By leveraging the spatial and multipath propagation characteristics of CSI and MIMO systems, fingerprinting enhances the accuracy and reliability of indoor localisation.

\subsection{Deep Learning for Position Estimation}

Deep learning has emerged as a powerful tool for enhancing indoor position estimation by leveraging the detailed signal characteristics captured by MIMO and CSI. By integrating deep learning techniques with the fingerprinting method, researchers can develop models that learn complex patterns and spatial relationships from the rich dataset of wireless signals, significantly improving the accuracy of indoor localisation systems. CNN and MLP are two types of neural networks effectively used in this context. CNNs are particularly well-suited for processing and extracting features from synthetic images, as they can learn spatial hierarchies and patterns within the data. MLPs are useful for combining these extracted features and performing precise position predictions~\cite{MLP}.

In~\cite{original}, the authors explored the integration of CSI and deep learning for indoor positioning within a MIMO system. A CNN was designed to estimate user positions using the fingerprint of the CSI data collected from 64 antennas over an indoor area of 2.5 by 2.5 meters. The study showed that the proposed system can achieve high accuracy in indoor localisation with the proposed method, reporting a mean error of under 6 cm in the estimations. The favourable outcomes obtained in these works aimed us to explore their use in the development of robust mobile robot localisation systems.

\subsection{Simulation Tools}

The process of developing (and validating) robotic localisation algorithms is characterised by its high cost and complexity. This point commonly reduces the range of test cases and limits the evaluation procedure. Robotic simulators are perfect for overcoming these problems, and they enable agile and powerful simulated test scenarios generation, obviating the costs and complexity associated with real-worlds experiments. One of the most interesting simulation engines is Webots, fully integrated with the Robot Operating System (ROS), and with an extensive set of simulated robots and world definitions. Webots programming capabilities are essential for its integration in higher-level applications used, for instance, to evaluate the performance of localisation solutions~\cite{ref_webots}.

\section{Localisation from CSI Readings} \label{sec:3}

\subsection{Data Preprocessing}

Our proposal starts with the preprocessing of CSI data, which is initially presented as the matrix of complex numbers $\boldsymbol{M}\in\mathbb{C}^{A\times S}$, where $A$ represents the number of antennas and $S$ denotes the number of subcarriers. The complex data is then converted to polar coordinates, resulting in modulus and argument (in radians) values. This transformation creates the dataset, with each measurement position represented by rows and the converted polar values forming the columns. Each row includes the X and Y positions as dependent variables of the CSI readings. Table~\ref{tab:tidy_data} shows the format of the dataset.

\begin{table}[htbp]
\caption{Example of the dataset format. \textit{A} is the antenna, \textit{S} is the subcarrier, $m$ is the modulus, $\phi$ is the argument, \textit{PosX} is the position \(X\), and \textit{PosY} is the position \(Y\).}
    \centering
    \begin{tabular}{ccccccc}
    \hline
        \textbf{A1S1-$m$} & \textbf{A1S1-$\phi$} & \textbf{A1S2-$m$} & \textbf{A1S2-$\phi$} & $\cdots$ & \textbf{PosX} & \textbf{PosY} \\ \hline
        0.234 & 0.643 & 0.275 & 0.631 & $\cdots$ & 302 & 2391 \\        
        $\cdots$ & $\cdots$ & $\cdots$ & $\cdots$ & $\cdots$ & $\cdots$ & $\cdots$ \\        
        0.180 & -1.989 & 0.152 & -1.961 & $\cdots$ & -1215 & 1221 \\        \hline
    \end{tabular}
    \label{tab:tidy_data}
\end{table}

\subsection{Images Generation From Data}

For the input data of the CNN, the TINTO tool is used to transform CSI readings into synthetic images. This process, detailed in~\cite{tinto2}, involves applying a dimensionality reduction algorithm to the CSI data, followed by translating and scaling the coordinates to generate 2D images that represent the locations of characteristic pixels. Furthermore, TINTO includes a blurring technique, which adds more ordered information to the images, improving the regression task in the CNN. In our proposal, all generated images have been created with dimensions of 35 x 35 pixels, using the blurring technique.

\subsection{Model Training}

\begin{figure}[ht]
    \centering
    \includegraphics[width=\linewidth]{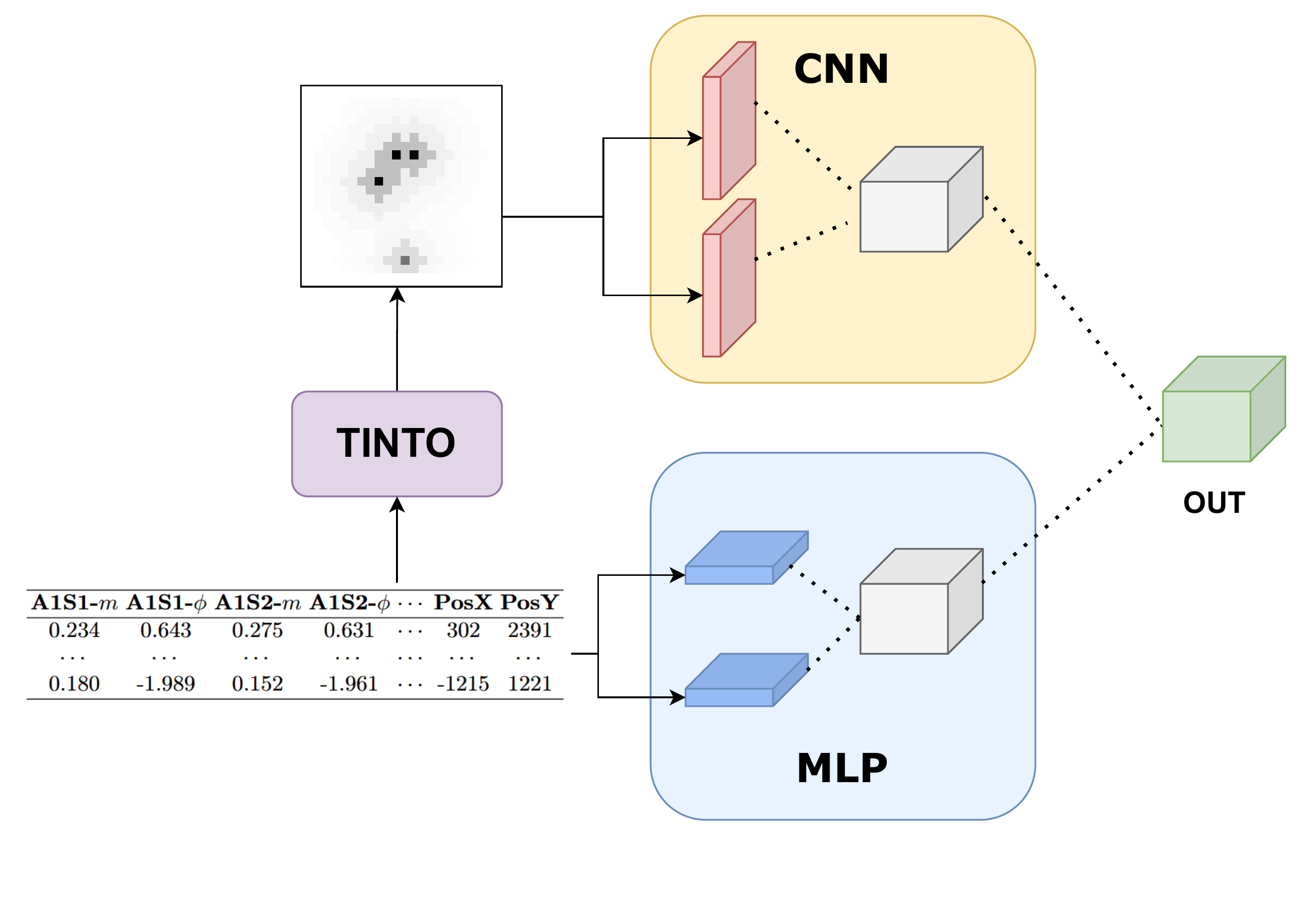}
    \caption{Simplified architecture of the Hybrid Neural Network.}
    \label{fig:architec}
\end{figure}

For our proposal, a HyNN has been developed. The HyNN model is capable of reading images on the CNN branch, and data on the MLP branch, enhancing the model ability to learn spatial feature. Each branch consists of two sub-branches. The CNN processes images with convolution, max pooling, and average pooling layers. The MLP processes the CSI data through dense layers with normalisation and dropout for regularisation. These branches are integrated through final dense and concatenation layers, leading to the output layer. Fig.~\ref{fig:architec} illustrates a simplified schematic of the HyNN architecture.

The dataset contains X and Y positions as target variables, which requires two models for each scenario: one to predict X and the other to predict Y. We used an 85\% train, 10\% validation, and 5\% test split for training, implemented with Keras and TensorFlow.

\subsection{Integration into a Simulation Procedure}\label{sec:3_D}

To evaluate our proposed localisation system in a robotics simulator, the following steps are necessary. By following these steps, we can assess the performance of the localization algorithm in a controlled, simulated environment:

\begin{enumerate}
\item Scenario reconstruction. Reconstruct the real scenario in the simulator, maintaining the Cartesian coordinates used in the real environment.

\item Positions of the route. Start the simulation to obtain the actual robot position along the pre-defined path.

\item CSI data of route positions. Associate the CSI readings from the preprocessed dataset with the route positions. For this association, calculate the mean of the CSI data from at least nine of the closest positions (with CSI data) to the actual robot position. This ensures that the data used for the evaluation closely approximate the actual conditions while preventing reuse of the training data.

\item Subsampling. Subsample the route positions to match the model prediction rate in a real-world scenario, adjusting for predictions/frames per second.

\item Noise application. Apply Gaussian noise to the route CSI readings, adjusting for modulus and argument ranges. Scenarios with no noise, low (10\%), medium (20\%) and high (30\%) noise levels are generated by setting the standard deviation as a percentage of each data range. Adding noise to the CSI data emulates the interference and fluctuations typical of real-world environments, ensuring the system is tested under conditions representative of practical, real-time operation.

\item Model prediction and evaluation. Launch the simulation, loading the models to predict the new CSI readings for each route position. This simulates real-world operation and allows for evaluating and comparing the average error between the actual position and the estimated position by the model.

\end{enumerate}

\section{Experiments and Results} \label{sec:4}

\subsection{Dataset} \label{sec:4_A}

\begin{figure}[ht]
    \centering
    \includegraphics[width=0.8\linewidth]{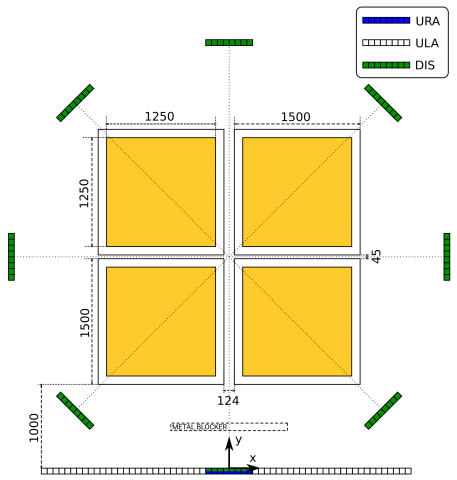}
    \caption{Scenario configuration. The antennas are spaced around a scenario where  yellow zones represent CSI readings positions. All measurements in the figure are in millimetres (Figure obtained from~\cite{dataset}).}
    \label{fig:scenario}
\end{figure}

For this research, the existing dataset created in~\cite{dataset} was used, which contains extensive raw CSI samples collected via 64 antennas. Using a pilot signal of 100 subcarriers, the CSI is determined, resulting in a matrix of [64 x 100] dimensions for each position. The dataset was created in a small, flat interior space, with dimensions illustrated in Fig.~\ref{fig:scenario}. CSI readings were obtained at 5-millimetre intervals across the entire area, yielding measurements at 252,004 positions.

Data was collected for three different antenna configurations: Uniform Rectangular Array (URA), Uniform Linear Array (ULA), and Distributed (DIS), as shown in Fig.~\ref{fig:scenario}. For our evaluation, we selected the ULA dataset to integrate into the robotics simulator. This choice is based on its practicality and ease of replication in real-world environments, making it a suitable example for this study. Moreover, we created additional datasets with varying numbers of antennas: 8, 16, 32, and 64. This allows us to study the accuracy based on the number of deployed antennas. Table~\ref{tab:hynn} shows the Mean Error (ME) results in function of the used antennas for ULA scenario. ME is calculated as:

\begin{equation*} ME (mm) = \mathbb{E} \{|p - \hat{p}|\},\tag{1}\end{equation*} 

\noindent where \(p\) is the real position and \(\hat{p}\) is the estimated position.

\begin{table}[htbp]
\caption{ME, in millimetres, of the ULA scenario for the HyNN models, in the test partition results.}\label{tab:hynn}
\centering
\begin{tabular}{|p{0.2\linewidth}|p{0.15\linewidth}|p{0.15\linewidth}|p{0.15\linewidth}|p{0.15\linewidth}|}
\hline
ULA             & 8 antennas & 16 antennas & 32 antennas & 64 antennas \\ \hline 
HyNN ME    & 164.98     & 137.37       & 104.53      & \textbf{49.66}       \\ \hline
\end{tabular}
\end{table}

The inclusion of a 64-antenna configuration allows for capturing a higher level of spatial diversity and multipath propagation characteristics inherent in MIMO systems, resulting in improved localisation accuracy. This configuration provides a theoretical reference for the model's performance under optimal conditions, while additional experiments with fewer antennas demonstrate the system's adaptability to scenarios with limited resources.

\subsection{Experimentation}

\begin{figure}[ht]
    \centering
    \includegraphics[width=\linewidth]{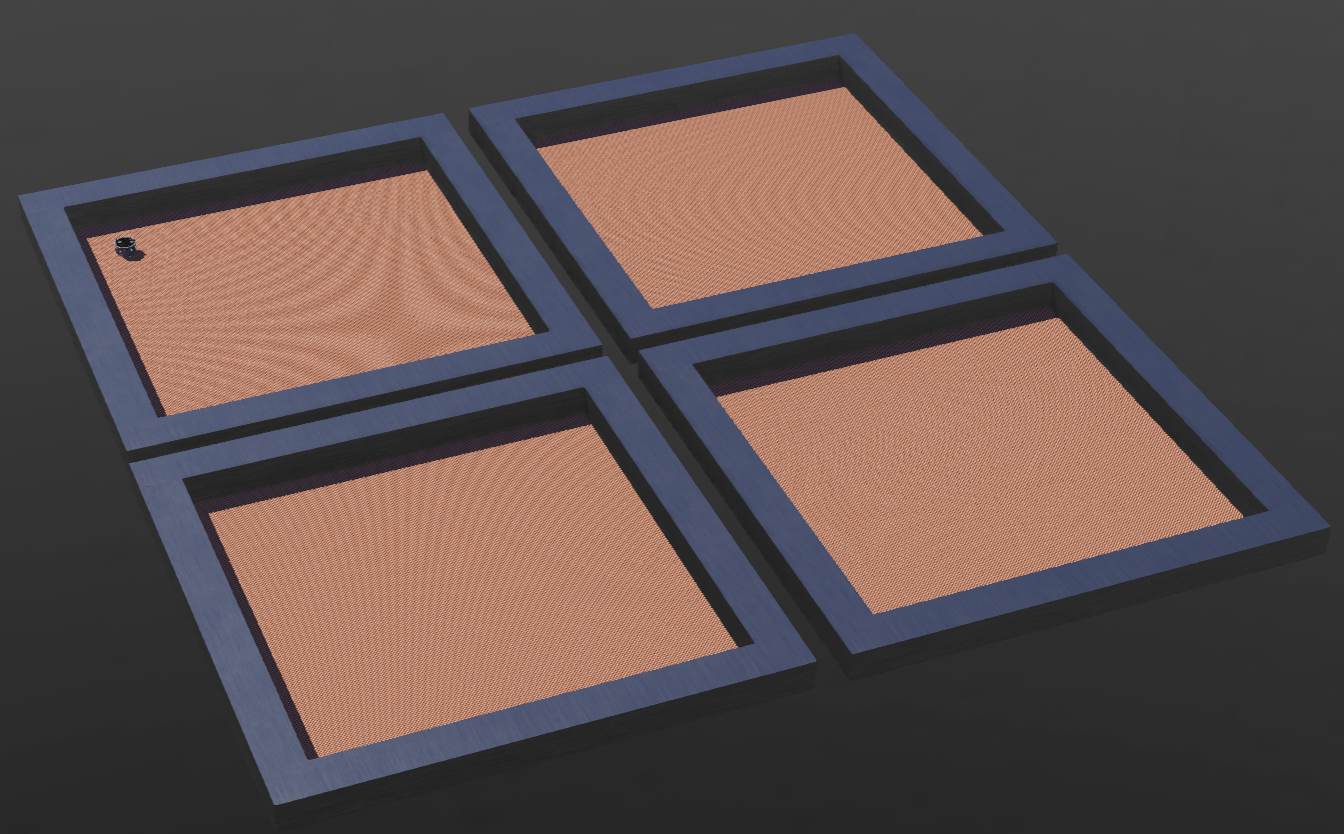}
    \caption{Reconstruction of the scenario in Webots.}
    \label{fig:scenario_webots}
\end{figure}

To proceed with the experimentation in Webots, the actual scenario was initially reconstructed in the simulator, adhering to the coordinates delineated in the dataset, as shown in Fig~\ref{fig:scenario_webots}. By employing the Webots Supervisor node, it is possible to determine the position of the robot in relation to the dataset (second step defined in Section~\ref{sec:3_D}). The robot selected for experimentation is the e-puck2\footnote{\url{https://www.cyberbotics.com/doc/guide/epuck?version=cyberbotics:R2019a}} robot, mainly due to its small size (7.5 cm in diameter), and its compatibility with Wi-Fi through a built-in microcontroller, which allows it to read CSI signals and use the developed localisation system. The efficacy of the proposed solution will be assessed based on its capacity to address various navigation problems, focusing on the system performance in estimating the robot position along a given route.

To improve and optimise the HyNN predictions, a simple implementation of a Kalman filter has been performed. By incorporating a two-dimensional Kalman filter, which state is represented by the X and Y position coordinates and the velocity in each axis, we can combine  the measurements from the HyNN with the state predictions of the filter. Although the state transition matrix is simplified to account only for the effect of velocity on position at each simulator time step, assuming constant velocity, the Kalman filter still provides valuable information about the robot velocity and helps to smooth out the estimations. This implementation allows us to predict future locations and behaviours based on the current and previous positions and velocities, contributing to the overall effectiveness of the localisation system. This implementation can be easily adapted to other scenarios and conditions, as it does not require performing a definition of the robot motion model.

Once the integration of our proposal to the dataset has been done (Section~\ref{sec:3_D}), we aim to demonstrate its effectiveness and reliability in practical applications by evaluating the system through specific test cases or experiments. The system will be evaluated in terms of:

\begin{itemize}
\item{\textbf{Number of antennas.} The number of antennas used to collect and measure the CSI data. The results obtained for 8, 16, 32 and 64 antennas will be studied.}
\item{\textbf{Noise level.} Indicates the amount of interference or randomness present in the data. The impact of noise on the CSI data will be evaluated, considering low, medium, high and no noise scenarios.}
\item{\textbf{Kalman filter.} Algorithm used to estimate the system state from noisy measurements, with the objective of reducing the error. In this case, the measurements are those predicted by HyNN. The results will be studied when applying or not applying the Kalman filter.}
\end{itemize}

The source code for the research is available in a public repository on GitHub\footnote{\url{https://github.com/Javierbj02/HyNN-Robot-Positioning}}.

\subsubsection{Prediction Times}

\begin{figure}[htbp]
    \centering
    \includegraphics[width=\linewidth]{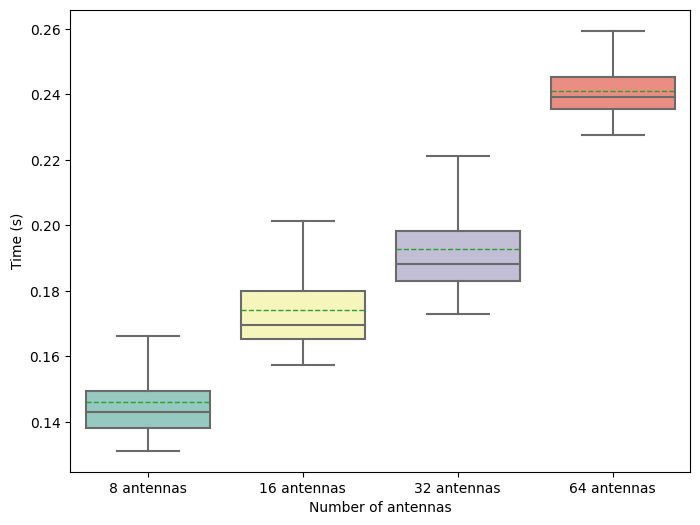}
    \caption{Box plot of the prediction times in seconds of the models as a function of the number of antennas.}
    \label{fig:times}
\end{figure}

Prior to initiating the experimental procedures, it is essential to measure the prediction times of the models in accordance with the number of antennas used. By analysing the prediction times, it will be possible to define the frames per seconds (FPS) at which to work, thus performing a sampling for each model corresponding to the prediction rate in real conditions. Fig.~\ref{fig:times} presents a box plot that provides a comprehensive illustration of the variation in HyNN model prediction time as a function of  the number of antennas used. Consequently, each box represents the distribution of prediction times for a specific number of antennas. It is evident that there is a direct correlation between the number of antennas and the prediction time. As the number of antennas increases, the prediction time also increases, which is to be expected given that the use of a larger number of antennas implies the processing of a greater quantity of data.

Following this analysis, it can be concluded that for datasets comprising 8, 16, 32, and 64 antennas, the models are capable of performing 5, 4, 4, and 3 predictions per second, respectively. These FPS will be used to sample the data for evaluation of the system, as outlined in Section~\ref{sec:3_D}.

\subsubsection{Experiment 1: Position Estimation during Uniform Motion.}

The objective of this experiment is to validate an entire procedure and to analyse whether the mean error ranges, documented in Sec.~\ref{sec:4_A}, are in line with those observed in the course of this experiment. For this purpose, the robot follows a straight-line path with a length of 1429 millimetres at a constant velocity.

\begin{figure}[htbp]
    \centering
    \includegraphics[width=0.8\linewidth]{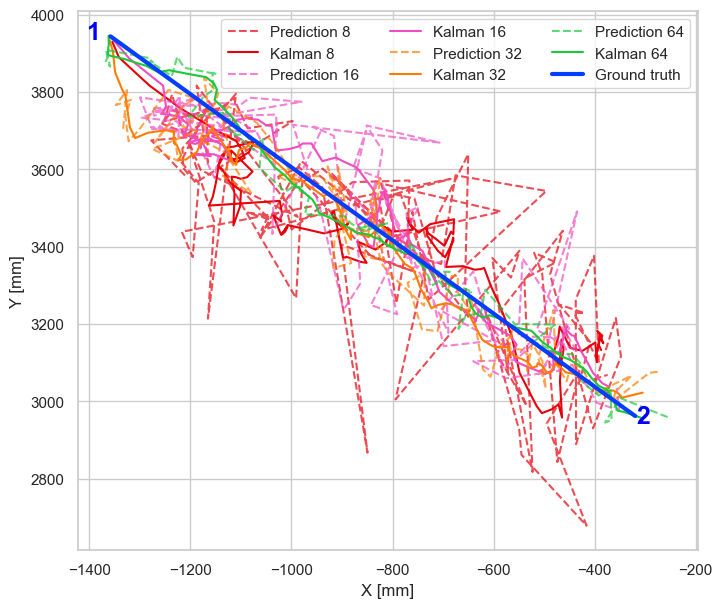}
    \caption{Results of predictions obtained in Experiment 1.}
    \label{fig:tc1}
\end{figure}

Table~\ref{tab:tc1} shows the Mean Error in millimetres obtained in the first experiment, for each noise level and number of antennas, with a comparison of the results with and without the Kalman filter. Fig.~\ref{fig:tc1} shows a plot of the results of the predictive models with and without the Kalman filter, with the mean across all noise levels plotted for each number of antennas. The dashed lines represent the HyNN predictions, while the solid lines illustrate the results of integrating the Kalman filter. The blue line depicts the actual trajectory of the robot, starting at point 1 and concluding at point 2.

\newcolumntype{C}[1]{>{\centering\let\newline\\\arraybackslash\hspace{0pt}}m{#1}}
\begin{table}[htbp]
\caption{Mean Error Results in millimetres obtained in Experiment 1.}
\centering
\begin{tabular}{C{2cm}C{2.3cm}|C{1.5cm}C{1.5cm}C{1.5cm}C{1.5cm}}
\hline
\multirow{2}{*}{Noise level} & \multirow{2}{*}{Kalman Filter} & \multicolumn{4}{c}{Number of antennas} \\
                        &     & 8 & 16 & 32 & 64 \\ \hline
\multirow{2}{*}{None}   & No  & 162.32     & 128.71      & 82.12      & 40.75      \\
                        & Yes & 124.64     & 95.76      & 59.12      & \textbf{31.90}      \\ \hline
\multirow{2}{*}{Low}    & No  & 179.73     & 133.99      & 80.89      & 37.68      \\
                        & Yes & 106.71     & 93.02      & 54.68      & \textbf{26.15}      \\ \hline
\multirow{2}{*}{Medium} & No  & 247.50     & 169.53      & 85.10      & 42.18      \\ 
                        & Yes & 126.00     & 109.15      & 62.02      & \textbf{24.81}      \\ \hline
\multirow{2}{*}{High}   & No  & 381.23     & 186.51      & 96.47      & 45.20      \\
                        & Yes & 272.25     & 108.90      & 76.46      & \textbf{25.80}      \\ \hline
\end{tabular}
\label{tab:tc1}
\end{table}

After this experiment, it can be stated that the results obtained are in alignment with those achieved during training. Furthermore, it can be concluded that the scenarios with 8 and 16 are insufficient and will not be included in the subsequent plots. However, they will be studied in the same way.

Although it is not the main objective of this experiment, we can highlight how the Kalman filter significantly improves all the results in a straight path at constant speed. This improvement is especially evident the higher the noise in the measurements, demonstrating its effectiveness in filtering noise, significantly increasing the robustness of the localisation algorithm.

\subsubsection{Experiment 2: Position Estimation during Obstacle Avoidance.}

The objective of this experiment is to evaluate the effectiveness of a state estimator that operates on time-series data, such as the Kalman filter, in order to improve the accuracy of the position estimation provided by the HyNN. In addition, it is intended to study the influence of noise on the results. To this end, a route of 1880 millimetres with obstacles is defined, being necessary to avoid them in order to progress.

\begin{figure}[htbp]
    \centering
    \includegraphics[width=0.7\linewidth]{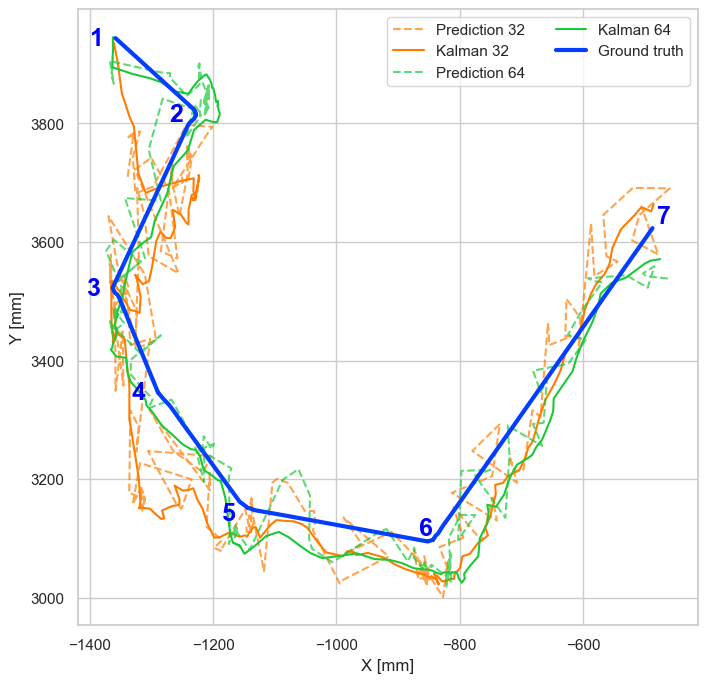}
    \caption{Results of predictions obtained in Experiment 2.}
    \label{fig:tc2}
\end{figure}

Table~\ref{tab:tc2} illustrates the Mean Error in millimetres obtained in Experiment 2 for each noise level and number of antennas, comparing also the use or not of the Kalman filter. Fig.~\ref{fig:tc2} depicts a comparison of the prediction outcomes with and without the Kalman filter. For the 32 and 64 antenna configurations, the averaged outcomes across all noise levels have been plotted. The dashed lines correspond to the HyNN predictions, whereas the solid lines illustrate the integrated Kalman filter results. The blue line represents the actual robot trajectory, beginning at point 1 and ending at point 7.

\begin{table}[htbp]
\caption{Mean Error Results in millimetres obtained in Experiment 2.}
\centering
\begin{tabular}{C{2cm}C{2.3cm}|C{1.5cm}C{1.5cm}C{1.5cm}C{1.5cm}}
\hline
\multirow{2}{*}{Noise level} & \multirow{2}{*}{Kalman Filter} & \multicolumn{4}{c}{Number of antennas} \\
                        &     & 8 & 16 & 32 & 64 \\ \hline
\multirow{2}{*}{None}   & No  & 150.27     & 114.05      & 83.10      & \textbf{42.91}      \\
                        & Yes & 131.29     & 109.45      & 83.37      & 49.91      \\ \hline
\multirow{2}{*}{Low}    & No  & 171.45     & 128.37      & 81.86      & \textbf{46.51}      \\
                        & Yes & 124.64     & 113.34      & 80.84      & 52.06      \\ \hline
\multirow{2}{*}{Medium} & No  & 317.36     & 169.10      & 79.68      & \textbf{48.66}      \\
                        & Yes & 230.48     & 125.45      & 77.19      & 51.37      \\ \hline
\multirow{2}{*}{High}   & No  & 376.77     & 221.48      & 91.70      & 57.88      \\
                        & Yes & 279.60     & 173.77      & 86.66      & \textbf{53.05}      \\ \hline
\end{tabular}
\label{tab:tc2}
\end{table}

Upon examination of the results, it becomes evident that in most cases the application of the Kalman filter significantly reduces the mean error, which becomes especially evident in the presence of elevated noise levels. This integration demonstrates its efficacy in environments with noise present, providing a more accurate estimation and enhancing the location algorithm's reliability in such contexts. Conversely, in scenarios where noise is minimal, the use of Kalman filter may even prove to be suboptimal, potentially due to the time required for stabilisation. Moreover, it can be observed that when 64 antennas are employed, thereby enabling the HyNN to perform more accurate predictions, the Kalman filter performs less well in the majority of cases, due to the simplified Kalman filter model.

While the efficacy of the Kalman filter could be enhanced by employing a parameterisation and motion model according to each robot, the objective has been to develop a straightforward implementation without defining specific motion models to facilitate adaptation to other problems. This explains why it performs better on a path such as that in Experiment 1, where the velocity remains constant throughout, whereas in Experiment 2 the robot adjusts its speed as it progresses to avoid obstacles.

\subsubsection{Experiment 3: Position Estimation in Robot Kidnapping Problem.}

In the context of robotics, the kidnapped robot problem refers to the loss of knowledge of a robot about its position due to a forced and unexpected displacement by external agents.

The aim of this experiment is to analyse the performance of the developed system in a scenario that can occur in real robots, such as the kidnapped problem, in order to measure the position recovery factor under these conditions. This experiment is similar to Experiment 2, since the robot has to avoid obstacles. However, at some point in the simulation, the robot is abruptly moved to a different position. The path length is 1779 millimetres.

\begin{figure}[htbp]
    \centering
    \includegraphics[width=0.8\linewidth]{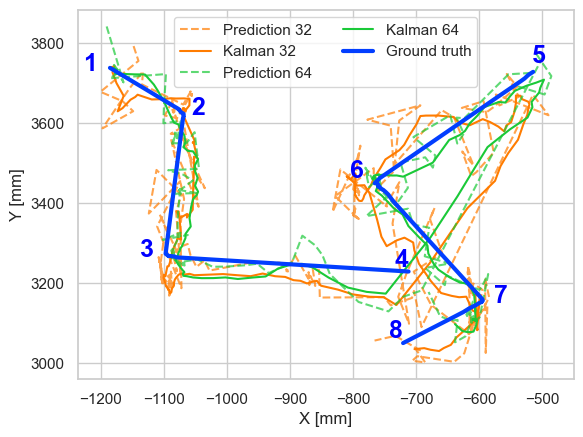}
    \caption{Results of predictions obtained in Experiment 3.}
    \label{fig:tc3}
\end{figure}

Table~\ref{tab:tc3} shows the Mean Error in millimetres obtained in Experiment 3, for each noise level and number of antennas, with a comparison of the results with and without the Kalman filter. Fig.~\ref{fig:tc3} depicts the predictive outcomes with and without the Kalman filter, with the results for the 32 and 64 antenna configurations averaged across all noise levels. The dashed lines correspond to the HyNN predictions, while the solid lines represent the results that integrate the Kalman filter. The blue line depicts the actual trajectory of the robot, which begins at point 1 and concludes at point 8. The jump between points 4 and 5 represents the kidnapping of the robot.

\begin{table}[htbp]
\caption{Mean Error Results in millimetres obtained in Experiment 3.}
\centering
\begin{tabular}{C{2cm}C{2.3cm}|C{1.5cm}C{1.5cm}C{1.5cm}C{1.5cm}}
\hline
\multirow{2}{*}{Noise level} & \multirow{2}{*}{Kalman Filter} & \multicolumn{4}{c}{Number of antennas} \\
                        &     & 8 & 16 & 32 & 64 \\ \hline
\multirow{2}{*}{None}   & No  & 156.40     & 112.08      & 61.94      & \textbf{35.56}      \\
                        & Yes & 146.55     & 109.69      & 65.43      & 49.44      \\ \hline
\multirow{2}{*}{Low}    & No  & 204.78     & 133.62      & 60.72      & \textbf{37.84}      \\
                        & Yes & 144.82     & 104.60      & 61.09      & 51.19      \\ \hline
\multirow{2}{*}{Medium} & No  & 360.67     & 166.89      & 66.14      & \textbf{41.46}      \\
                        & Yes & 226.89     & 131.77      & 59.42      & 53.19      \\ \hline
\multirow{2}{*}{High}   & No  & 387.21     & 198.17      & 81.60      & \textbf{46.75}      \\
                        & Yes & 236.75     & 121.70      & 66.76      & 51.01      \\ \hline
\end{tabular}
\label{tab:tc3}
\end{table}

In this experiment, it is observed that when the initial estimates are accurate (64-antenna setup), the Kalman filter demonstrates a systematic decline in performance. In contrast, in the presence of substantial noise, with an initial error exceeding 70 millimetres, the Kalman filter demonstrates a consistent improvement. Furthermore, it is crucial to note that in a kidnapping situation, the robot motion becomes highly nonlinear. This renders the Kalman filter unable to accurately estimate the new robot position after the kidnapping, resulting in a prolonged convergence period to the new actual position. This phenomenon elucidates the inferior performance observed in the 32- and 64-antenna scenarios when employing the Kalman filter, where the HyNN already obtains high prediction accuracy.

\section{Conclusions and Future Work} \label{sec:5}

We have presented a procedure to integrate a position estimation algorithm, fed by CSI data and based on a Hybrid Neural Network approach, into a higher-level solution that includes a robotic operating system, simulation capabilities, and the Kalman filter localisation algorithm. This proposal allows for flexible design and development of test cases, enabling in-depth analysis of the position estimation algorithm. The cornerstone of the proposal is the synthetic generation of CSI readings based on an existing dataset. Based on the obtained results, we can conclude that the position estimator is suitable for real-world robotic scenarios, even in the presence of noise. Specifically, using 32 or 64 antennas ensures an average error of approximately 10 cm or less (135\% of the robot’s diameter). 

Future work will focus on validating the system in real-world environments to address the limitations of simulation-based evaluation and assess robustness under practical conditions. This will include testing on physical robots in diverse indoor scenarios and exploring trade-offs between cost and accuracy by analysing configurations with different datasets generated from wireless readings. We plan to integrate different state estimators, such as the Monte Carlo localisation algorithm, for dynamic contexts.

\sloppy
\section*{Acknowledgements}
This work has been funded by the project SBPLY/21/180225/000062 funded by the Government of Castilla-La Mancha and “ERDF A way of making Europe”. This work has also been supported by Universidad de Castilla-La Mancha and “ERDF A way of making Europe” under project 2023-GRIN-34437, and grant PID2022-137344OB-C33 funded by MICIU/AEI/10.13039/501100011033 and by "ERDF/EU".

\end{document}